\pdfoutput=1

\documentclass[11pt]{article}

\usepackage[]{acl}

\usepackage{times}
\usepackage{latexsym}

\usepackage[T1]{fontenc}

\usepackage[utf8]{inputenc}

\usepackage{microtype}

\usepackage{amsmath}
\usepackage{amsfonts}
\usepackage{lipsum} 
\usepackage{algorithm}
\usepackage{setspace}
\usepackage{graphicx}
\usepackage[moderate]{savetrees}

\usepackage{xcolor}
\definecolor{bluenode}{HTML}{00a7db}
\definecolor{rednode}{HTML}{ea4e00}
\definecolor{purplenode}{HTML}{9d00d7}
\definecolor{graynode}{HTML}{898989}
\definecolor{ForestGreen}{RGB}{34,139,34}
\usepackage{multirow}
\usepackage{makecell}
\usepackage{arydshln}
\usepackage{booktabs}

\usepackage{amsmath,amsfonts,bm}

\def\eqref#1{equation~\ref{#1}}

\def\1{\bm{1}}

\def\vh{{\bm{h}}}

\def\vw{{\bm{w}}}

\DeclareMathAlphabet{\mathsfit}{\encodingdefault}{\sfdefault}{m}{sl}
\SetMathAlphabet{\mathsfit}{bold}{\encodingdefault}{\sfdefault}{bx}{n}

\def\gE{{\mathcal{E}}}

\def\gG{{\mathcal{G}}}

\def\gN{{\mathcal{N}}}

\def\gR{{\mathcal{R}}}

\def\gV{{\mathcal{V}}}

\usepackage{mathtools}
\usepackage{xspace}
\newcommand{\eg}{\textit{e.g.}}
\newcommand{\ie}{\textit{i.e.}}
\newcommand{\heading}[1]{\vspace{1mm}\noindent\textbf{#1}~~}
\newcommand{\methodname}{QA-GNN\xspace}

\newcommand{\schema}{\gG_\text{sub}}
\newcommand{\snode}{\gV_\text{sub}}
\newcommand{\sedge}{\gE_\text{sub}}
\newcommand{\casnode}{\gV_\text{W}}
\newcommand{\casedge}{\gE_\text{W}}
\newcommand{\cas}{\gG_\text{W}}
\newcommand{\qnode}{\gV_q}
\newcommand{\anode}{\gV_a}
\newcommand{\qanode}{\gV_{q,a}}
\newcommand{\fembed}{f_\text{enc}}
\newcommand{\fscore}{f_\text{head}}
\def\nodetext#1{\texttt{text}(#1)}

\renewcommand\ttdefault{cmtt}
\hypersetup{
  colorlinks   = true, 
  urlcolor     = ForestGreen, 
  linkcolor    = darkblue, 
  citecolor   = darkblue 
}

\title{\methodname: Reasoning with Language Models and Knowledge Graphs \\ for Question Answering}

\author{Michihiro Yasunaga, ~~ Hongyu Ren, ~~ 
Antoine Bosselut\\
{\bf Percy Liang, ~~ Jure Leskovec}\\
Stanford University\\
\scalebox{0.87}[0.9]{{\tt \{myasu,hyren,antoineb,pliang,jure\}@cs.stanford.edu}}}

\begin{document}
\setlength{\abovedisplayskip}{6pt}
\setlength{\belowdisplayskip}{6pt}

\maketitle

\begin{abstract} 
The problem of answering questions using knowledge from pre-trained language models (LMs) and knowledge graphs (KGs) presents two challenges:
given a QA context (question and answer choice), methods need to (i) identify relevant knowledge from large KGs, and (ii) perform joint reasoning over the QA context and KG.
In this work, we propose a new model, \textit{\methodname}, which addresses the above challenges through two key innovations:
(i) relevance scoring, where we use LMs to estimate the importance of KG nodes relative to the given QA context, 
and (ii) joint reasoning, where we connect the QA context and KG to form a joint graph, and mutually update their representations through graph neural networks. 
We evaluate our model on QA benchmarks in the commonsense (CommonsenseQA, OpenBookQA) and biomedical (MedQA-USMLE) domains. \methodname outperforms existing LM and LM+KG models, and exhibits capabilities to perform interpretable and structured reasoning, \eg, correctly handling negation in questions. {Our code and data are available at \url{https://github.com/michiyasunaga/qagnn}.}
\end{abstract}
\section{Introduction}\label{sec:intro}

Question answering systems must be able to access relevant knowledge and reason over it. Typically, knowledge can be implicitly encoded in large language models (LMs) pre-trained on unstructured text  \cite{petroni2019language,Bosselut2019COMETCT}, or explicitly represented in structured knowledge graphs (KGs), such as Freebase \cite{bollacker2008freebase} and ConceptNet \cite{speer2016conceptnet}, where entities are represented as nodes and relations between them as edges.
Recently, pre-trained LMs have demonstrated remarkable success in many question answering tasks \cite{liu2019roberta,t5}. However, while LMs have a broad coverage of knowledge, they do not empirically perform well on structured reasoning (\eg, handling negation) \cite{kassner2020negated}. On the other hand, KGs are more suited for structured reasoning \cite{ren2020query2box,ren2020beta} and enable explainable predictions \eg, by providing reasoning paths \cite{lin2019kagnet}, but may lack coverage and be noisy \cite{bordes2013translating,guu2015traversing}. How to reason effectively with both sources of knowledge remains an important open problem.

\begin{figure}[!t]
    \centering
    \includegraphics[width=0.45\textwidth]{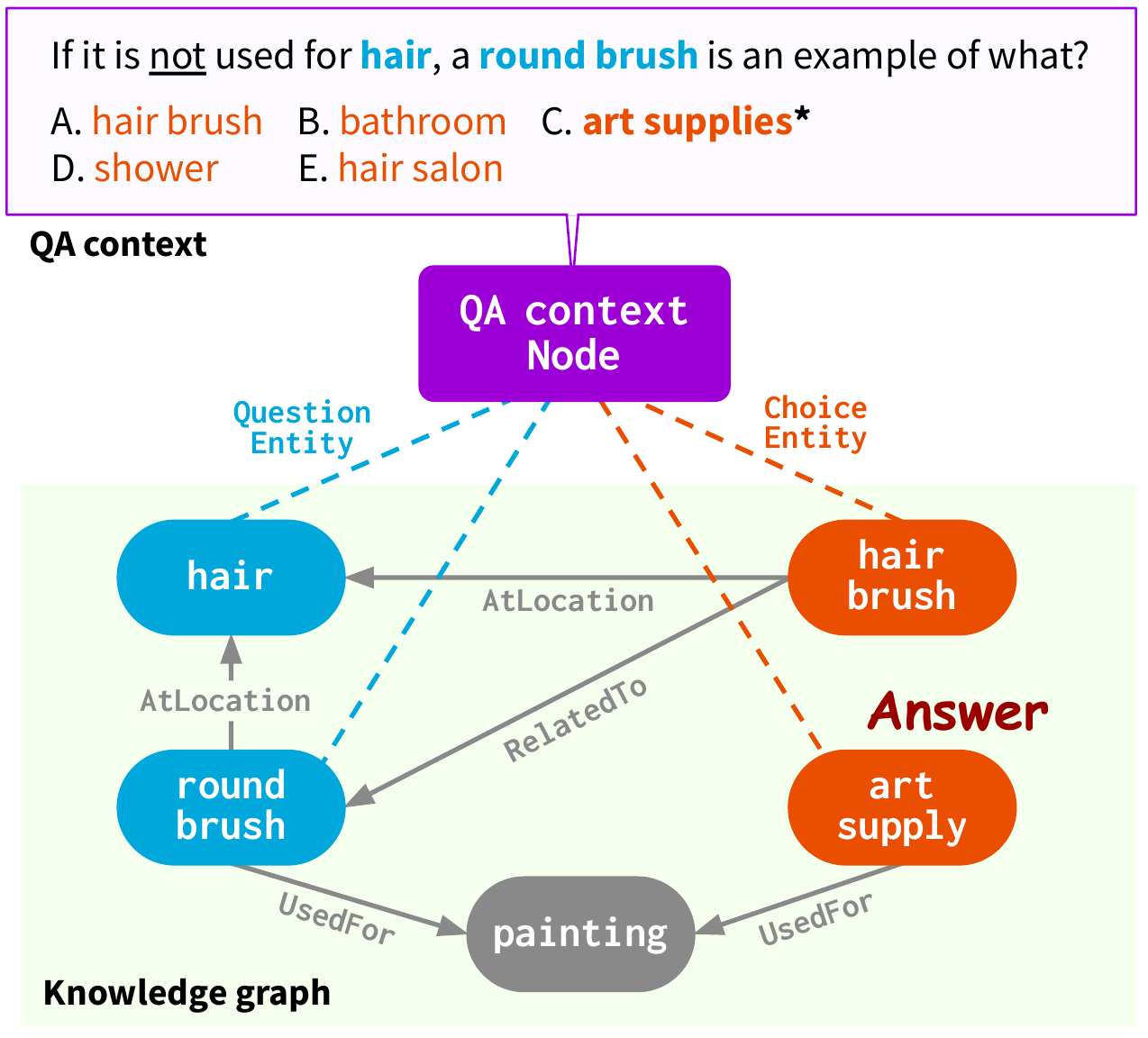}\vspace{-2mm}
    \caption{Given the QA context (question and answer choice; purple box), we aim to derive the answer by performing joint reasoning over the language and the knowledge graph (green box). 
    }
  \label{fig:task}
\end{figure}

\begin{figure*}[!th]
    \vspace{-2mm}
    \centering 
    \includegraphics[width=0.98\textwidth]{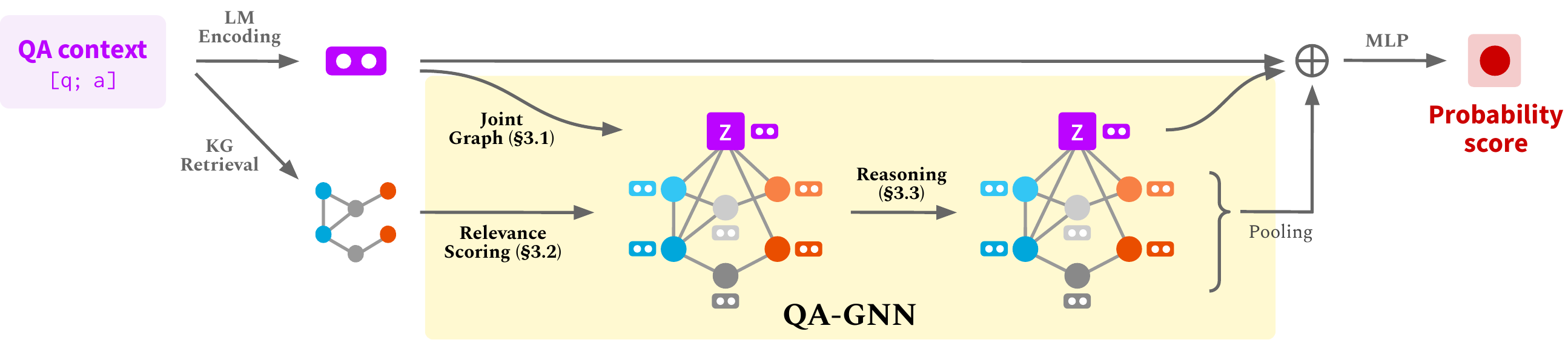}
    \vspace{-1mm}
    \caption{Overview of our approach. Given a QA context ($z$), 
    we connect it with the retrieved KG to form a joint graph (\textit{working graph}; \S \ref{sec:method-joint-graph}), compute the relevance of each KG node conditioned on $z$ (\S \ref{sec:method-contextualization}; node shading indicates the relevance score), and perform reasoning on the working graph (\S \ref{sec:method-gnn}). 
    }\label{fig:overview}
\end{figure*}

Combining LMs and KGs for reasoning (henceforth, LM+KG) presents two challenges: given a QA context (\eg, question and answer choices; Figure 1 purple box), methods need to (i) identify informative knowledge from a large KG (green box); and (ii) capture the nuance of the QA context and the structure of the KGs to perform joint reasoning over these two sources of information. 
Previous works \cite{bao2016constraint,sun2018open,lin2019kagnet} retrieve a subgraph from the KG by taking \textit{topic entities} (KG entities mentioned in the given QA context) and their few-hop neighbors. However, this introduces many entity nodes that are semantically irrelevant to the QA context, especially when the number of topic entities or hops increases. 
Additionally, existing LM+KG methods for reasoning \cite{lin2019kagnet,wang2019improving,feng2020scalable,lv2020graph} treat the QA context and KG as two separate modalities. They individually apply LMs to the QA context and graph neural networks (GNNs) to the KG, and do not mutually update or unify their representations. This separation might limit their capability to perform structured reasoning, \eg, handling negation.

Here we propose \textit{\methodname}, an end-to-end LM+KG model for question answering that addresses the above two challenges. We first encode the QA context using an LM, and retrieve a KG subgraph following prior works \cite{feng2020scalable}. 
Our \methodname has two key insights: (i) \textbf{Relevance scoring}: Since the KG subgraph consists of all few-hop neighbors of the topic entities, some entity nodes are more relevant than others with respect to the given QA context. We hence propose KG node relevance scoring: 
we score each entity on the KG subgraph by concatenating the entity with the QA context and calculating the likelihood using a pre-trained LM. This presents a general framework to weight information on the KG; (ii) \textbf{Joint reasoning}: We design a joint graph representation of the QA context and KG, where we explicitly view the QA context as an additional node (\textit{QA context node}) and connect it to the topic entities in the KG subgraph as shown in Figure 1. 
This joint graph, which we term the \textit{working graph}, unifies the two modalities into one graph.
We then augment the feature of each node with the relevance score, and design a new attention-based GNN module for reasoning.
Our joint reasoning algorithm on the working graph simultaneously updates the representation of both the KG entities and the QA context node, bridging the gap between the two sources of information.

We evaluate \methodname on three question answering datasets that require reasoning with knowledge: \textit{CommonsenseQA} \cite{talmor2018commonsenseqa} and \textit{OpenBookQA} \cite{obqa} in the commonsense domain (using the \textit{ConceptNet} KG), and \textit{MedQA-USMLE} \cite{jin2021disease} in the biomedical domain (using the UMLS and DrugBank KGs). \methodname outperforms strong fine-tuned LM baselines as well as the existing best LM+KG model (with the same LM) by 4.7\% and 2.3\% respectively. In particular, \methodname exhibits improved performance on some forms of structured
reasoning (\eg, correctly handling negation and entity substitution in questions):
it achieves 4.6\% improvement over fine-tuned LMs on questions with negation, while existing LM+KG models are +0.6\% over fine-tuned LMs. 
We also show that one can extract reasoning processes from \methodname in the form of general KG subgraphs, not just paths \cite{lin2019kagnet}, suggesting a general method for explaining model predictions.
\section{Problem statement}\label{sec:setup}
\renewcommand\ttdefault{cmtt}
We aim to answer natural language questions using knowledge from a pre-trained LM and a structured KG.
We use the term language model broadly to be any composition of two functions, $\fscore(\fembed(\mathbf{x}))$, where $f_{\text{enc}}$, the encoder, maps a textual input $\mathbf{x}$ to a contextualized vector representation $\mathbf{h}^\text{LM}$, and $f_{\text{head}}$ uses this representation to perform a desired task (which we discuss in \S\ref{sec:method-contextualization}). 
In this work, we specifically use masked language models (\eg, RoBERTa) as $\fembed$, and let $\mathbf{h}^\text{LM}$ denote the output representation of a \texttt{[CLS]} token that is prepended to the input sequence $\mathbf{x}$, unless otherwise noted.
We define the knowledge graph as a multi-relational graph $\gG=(\gV, \gE)$.
Here $\gV$ is the set of entity nodes in the KG; $\gE \subseteq$ $\gV \times \gR \times \gV$ is the set of edges that connect nodes in $\gV$, where $\gR$ represents a set of relation types.

Given a question $q$ and an answer choice $a \in \mathcal{C}$, 
we follow prior work \cite{lin2019kagnet} to 
link the entities mentioned in the question and answer choice to the given KG $\gG$. 
We denote $\qnode \subseteq \gV$ and $\anode \subseteq \gV$ as the set of KG entities mentioned in the question (\textit{question entities}; \textcolor{bluenode}{\textbf{blue}} entities in Figure\! 1) and answer choice (\textit{answer choice entities}; \textcolor{rednode}{\textbf{red}} entities in Figure\! 1), respectively, and use $\qanode\coloneqq \qnode \cup \anode$ to denote all the entities that appear in either the question or answer choice, which we call \textit{topic entities}. We then extract a subgraph from $\gG$ for a question-choice pair, $\schema^{q,a}=(\snode^{q,a},\sedge^{q,a})$,\footnote{We remove the superscript $q,a$ if there is no ambiguity.\vspace{-0mm}} which comprises all nodes on the $k$-hop paths between nodes in $\qanode$.

\begin{figure*}[!th]
    \vspace{-2mm}
    \centering 
    \includegraphics[width=0.95\textwidth]{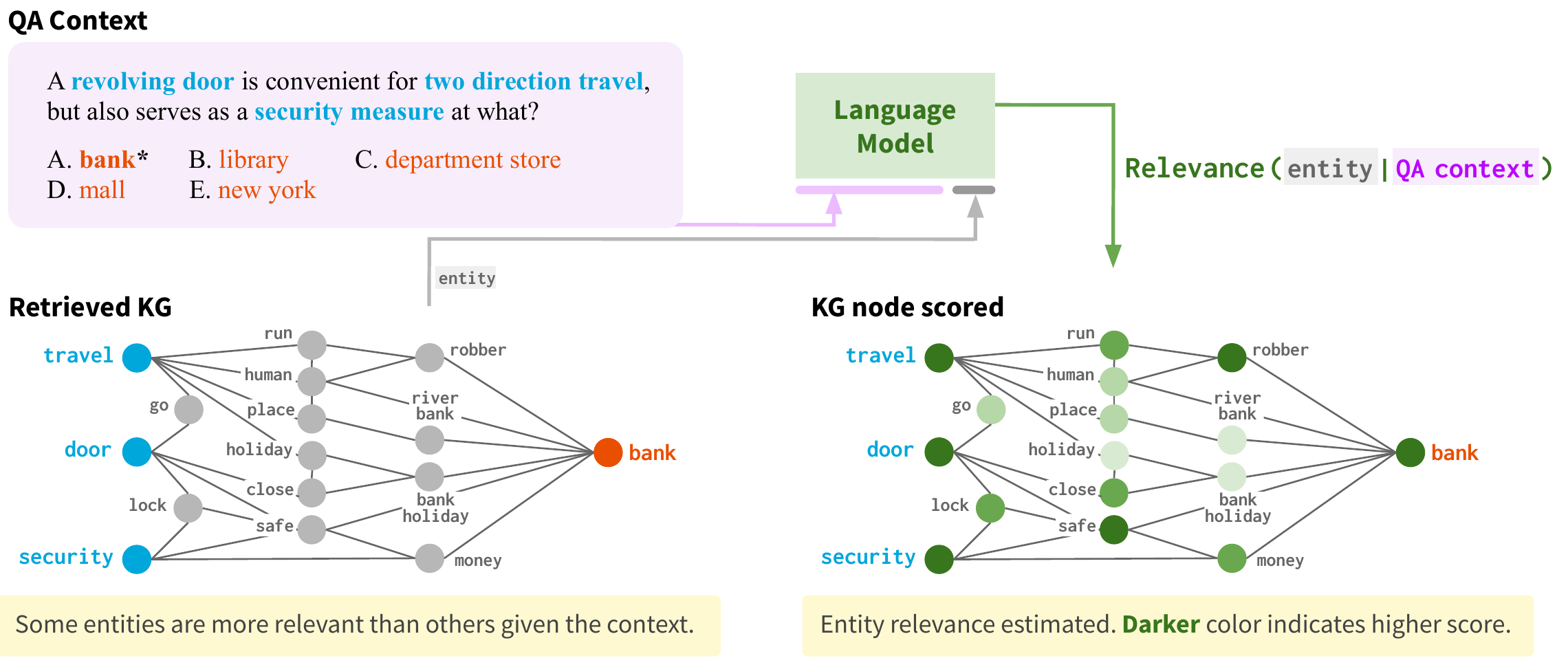}
    \vspace{-2mm}
    \caption{Relevance scoring of the retrieved KG: we use a pre-trained LM to calculate the relevance of each KG entity node conditioned on the QA context (\S \ref{sec:method-contextualization}).
    }
\label{fig:contextualization}
\end{figure*}

\section{Approach: \methodname}\label{sec:method}

As shown in Figure \ref{fig:overview},
given a question and an answer choice $a$, we concatenate them to get the \textit{QA context} $[q;~a]$.
To reason over a given QA context using knowledge from both the LM and the KG, \methodname works as follows.
First, we use the LM to obtain a representation for the QA context, and retrieve the subgraph $\schema$ from the KG.
Then we introduce a \textit{QA context node} $z$ that represents the QA context, and connect $z$ to the topic entities $\qanode$ so that we have a joint graph over the two sources of knowledge, which we term the \textit{working graph}, $\cas$ (\S \ref{sec:method-joint-graph}).
To adaptively capture the relationship between the QA context node and each of the other nodes in $\cas$, we calculate a relevance score for each pair using the LM, and use this score as an additional feature for each node (\S \ref{sec:method-contextualization}).
We then propose an attention-based GNN module that performs message passing on the $\cas$ for multiple rounds (\S \ref{sec:method-gnn}). We make the final prediction using the LM representation, QA context node representation and a pooled working graph representation (\S \ref{sec:method-inference}).

We also discuss the computational complexity of our model (\S \ref{sec:method-comp}), and why our model uses a GNN for question answering tasks (\S \ref{sec:method-why-gnn}).

\subsection{Joint graph representation}
\label{sec:method-joint-graph}

To design a joint reasoning space for the two sources of knowledge, we explicitly connect them in a common graph structure. We introduce a new QA context node $z$ which represents the QA context,
and connect $z$ to each topic entity in $\qanode$ on the KG subgraph $\schema$ using two new relation types $r_{z,q}$ and $r_{z,a}$.
These relation types capture the relationship between the QA context and the relevant entities in the KG, depending on whether the entity is found in the question portion or the answer portion of the QA context.
Since this joint graph intuitively provides a reasoning space (working memory) over the QA context and KG, we term it \textit{\textbf{working graph}} $\cas=(\casnode,\casedge)$, where $\casnode=\snode\cup\{z\}$ and $\casedge=\sedge\cup\{(z,r_{z,q},v) \mid v\in\qnode\}\cup\{(z,r_{z,a},v) \mid v\in\anode\}$.

Each node in $\cas$ is associated with one of the four types: $\mathcal{T}= \{\textbf{Z}, \textbf{Q}, \textbf{A}, \textbf{O}\}$, each indicating the context node $z$, nodes in $\qnode$, nodes in $\anode$, and other nodes, respectively (corresponding to the node color, \textcolor{purplenode}{\textbf{purple}}, \textcolor{bluenode}{\textbf{blue}}, \textcolor{rednode}{\textbf{red}}, \textcolor{graynode}{\textbf{gray}} in Figure \!1 and 2).
We denote the text of the context node $z$ (QA context) and KG node $v\in\snode$ (entity name) as $\nodetext{z}$ and $\nodetext{v}$.

We initialize the node embedding of $z$ by the LM representation of the QA context
($\boldsymbol{z}^\text{LM}$ $=$ $\fembed (\nodetext{z})$),
and each node on $\schema$ by its entity embedding (\S \ref{sec:experiment-kg}).
In the subsequent sections, we will reason over the working graph to score a given (question, answer choice) pair.

\subsection{KG node relevance scoring}
\label{sec:method-contextualization}
\renewcommand\ttdefault{cmtt}

Many nodes on the KG subgraph $\schema$ (\ie, those heuristically retrieved from the KG) can be irrelevant under the current QA context. As an example shown in Figure \ref{fig:contextualization}, the retrieved KG subgraph $\schema$ with few-hop neighbors of the $\qanode$ may include nodes that are uninformative for the reasoning process, \eg, nodes ``holiday'' and ``river bank'' are off-topic; ``human'' and ``place'' are generic.
These irrelevant nodes may result in overfitting or introduce unnecessary difficulty in reasoning, an issue especially when $\qanode$ is large. For instance, we empirically find that using the \emph{ConceptNet} KG \cite{speer2016conceptnet}, we will retrieve a KG with $|\snode|>400$ nodes on average if we consider 3-hop neighbors.

In response, we propose node relevance scoring, where we use the pre-trained language model to score the relevance of each KG node $v\in\snode$ conditioned on the QA context.
For each node $v$, we concatenate the entity $\nodetext{v}$ with the QA context $\nodetext{z}$ and compute the \emph{relevance score}:
\begin{align}
    \rho_v=\fscore(\fembed([\nodetext{z};~ \nodetext{v}])),\label{eq:relevance}
\end{align}
where $f_\text{head} \circ f_\text{enc}$ denotes
the probability of $\nodetext{v}$ computed by the LM.
This relevance score $\rho_v$ captures the importance of each KG node relative to the given QA context, which is used for reasoning or pruning the working graph $\cas$.

\subsection{GNN architecture}
\label{sec:method-gnn}

To perform reasoning on the working graph $\cas$, our GNN module builds on the graph attention framework (GAT) \citep{velikovi2017graph},
which induces node representations via iterative message passing between neighbors on the graph.
Specifically, in a $L$-layer \methodname, for each layer, we update the representation $\boldsymbol{h}_t^{(\ell)}\in\mathbb{R}^D$ of each node $t\in \casnode$ by
\begin{align}
    \boldsymbol{h}_{t}^{(\ell+1)}={f}_{n}
    \Biggl(\sum_{s \in \mathcal{N}_{t} \cup \{t\} } \alpha_{st} \boldsymbol{m}_{s t}\Biggr)
    +\boldsymbol{h}_{t}^{(\ell)},
\end{align}
where $\gN_t$ represents the neighborhood of node $t$, $\boldsymbol{m}_{s t} \in \mathbb{R}^D$ denotes the message from each neighbor node $s$ to $t$, and $\alpha_{st}$ is an attention weight that scales each message $\boldsymbol{m}_{s t}$ from $s$ to $t$. The sum of the messages is then passed through a 2-layer MLP,
${f}_{n}$: $\mathbb{R}^{D} \rightarrow \mathbb{R}^{D}$, with batch normalization \citep{ioffe2015batch}.
For each node $t\in\casnode$, we set $\boldsymbol{h}^{(0)}_t$ using a linear transformation $f_h$ that maps its initial node embedding (described in \S \ref{sec:method-joint-graph}) to $\mathbb{R}^D$.
Crucially, as our GNN message passing operates on the working graph, it will jointly leverage and update the representation of the QA context and KG.

\noindent
We further propose an expressive message ($\boldsymbol{m}_{st}$) and attention ($\alpha_{st}$) computation below.

\paragraph{Node type \& relation-aware message.}
\renewcommand\ttdefault{cmtt}
As $\cas$ is a multi-relational graph, the
message passed from a source node to the target node should capture their relationship, \ie, relation type of the edge and source/target node types.
To this end, we first obtain the type embedding $\boldsymbol{u}_{t}$ of each node $t$, as well as the relation embedding $\boldsymbol{r}_{st}$ from node $s$ to node $t$ by
\begin{align}
    \boldsymbol{u}_t  = {f}_{u} (\texttt{u}_t),~~~~
    \boldsymbol{r}_{st}  = {f}_r(\texttt{e}_{st},~ \texttt{u}_{s},~ \texttt{u}_{t}),
\end{align}
where $\texttt{u}_s, \texttt{u}_t \in \{0,1\}^{|\mathcal{T}|}$
are one-hot vectors indicating the node types of $s$ and $t$,
$\texttt{e}_{st} \in \{0,1\}^{|\mathcal{R}|}$ is a one-hot vector indicating the relation type of edge ($s,t$),
${f}_{u}$: $\mathbb{R}^{|\mathcal{T}|} \rightarrow \mathbb{R}^{D/2}$ is a linear transformation,
and
${f}_{r}$: $\mathbb{R}^{|\mathcal{R}|+2|\mathcal{T}|} \rightarrow \mathbb{R}^{D}$ is a 2-layer MLP.
We then compute the message from $s$ to $t$ as
\begin{align}
    \boldsymbol{m}_{st} &= {f}_{m}(\boldsymbol{h}_{s}^{(\ell)},~ \boldsymbol{u}_{s},~ \boldsymbol{r}_{s t}),
\end{align}
where ${f}_m$: $\mathbb{R}^{2.5D} \rightarrow \mathbb{R}^{D}$ is a linear transformation.

\paragraph{Node type, relation, and score-aware attention.}
Attention captures the strength of association between two nodes, which
is ideally informed by their node types, relations and node relevance scores.

\noindent
We first embed the relevance score of each node $t$ by
\begin{align}
    \boldsymbol{\rho}_t & = {f}_\rho (\rho_t),
\end{align}
where ${f}_{\rho}$: $\mathbb{R} \rightarrow \mathbb{R}^{D/2}$ is an MLP.
To compute the attention weight $\alpha_{st}$ from node $s$ to node $t$, we obtain the query and key vectors $\boldsymbol{q}$, $\boldsymbol{k}$ by
\begin{align}
    \boldsymbol{q}_s &= {f}_{q}(\boldsymbol{h}_{s}^{(\ell)},~ \boldsymbol{u}_{s},~ \boldsymbol{\rho}_{s}),\\
    \boldsymbol{k}_{t} &= {f}_{k}(\boldsymbol{h}_{t}^{(\ell)},~ \boldsymbol{u}_{t},~ \boldsymbol{\rho}_{t},~ \boldsymbol{r}_{s t}),
\end{align}
where ${f}_{q}$: $\mathbb{R}^{2D} \rightarrow \mathbb{R}^{D}$ and ${f}_{k}$: $\mathbb{R}^{3D} \rightarrow \mathbb{R}^{D}$ are linear transformations.
The attention weight is then
\begin{align}
    \alpha_{st} =\frac{\exp(\gamma_{st})}{\sum_{t' \in \mathcal{N}_{s} \cup \{s\}} \exp (\gamma_{st'})}, ~~~\gamma_{st} = \frac{\boldsymbol{q}_s^{\top} \boldsymbol{k}_{t}}{\sqrt{D}}.
\end{align}

\subsection{Inference \& Learning}
\label{sec:method-inference}
Given a question $q$ and an answer choice $a$, we use the information from both the QA context and the KG to calculate the probability of it being the answer $p(a \mid q)\propto \exp(\text{MLP}(\boldsymbol{z}^{\text{LM}},~ \boldsymbol{z}^{\text{GNN}},~ \boldsymbol{g}))$, where $\boldsymbol{z}^{\text{GNN}} =\vh^{(L)}_z$ and $\boldsymbol{g}$
denotes the pooling of $\{\vh^{(L)}_v \mid v \in \mathcal{V}_\text{sub}\}$.
In the training data, each question has a set of answer choices with one correct choice. We optimize the model (both the LM and GNN components end-to-end) using the cross entropy loss.

\subsection{Computation complexity}
\label{sec:method-comp}
We analyze the time and space complexity of our model and compare with prior works, KagNet \cite{lin2019kagnet} and MHGRN \cite{feng2020scalable} in Table \ref{tab:complexity}.
As we handle edges of different relation types using different edge embeddings instead of designing an independent graph networks for each relation as in RGCN \cite{schlichtkrull2018modeling} or MHGRN, the time complexity of our method is constant with respect to the number of relations and linear with respect to the number of nodes. We achieve the same space complexity as MHGRN \cite{feng2020scalable}.

\begin{table}[t]
\vspace{-0.0cm}
    \centering
     \scalebox{0.8}{
     \begin{tabular}{lcc}
      \toprule
      \textbf{Model} & \textbf{Time}  & \textbf{Space}  \\
      \midrule
      \multicolumn{3}{c}{\textit{$\mathcal{G}$ is a dense graph}} \\
      \midrule
      $L$-hop KagNet & \!\!$\mathcal{O}\left(|\gR|^L |\gV|^{L+1}L \right)$\!\! & \!\!$\mathcal{O}\left(|\gR|^L |\gV|^{L+1}L \right)$\!\! \\
      $L$-hop MHGRN & $\mathcal{O}\left(|\gR|^2|\gV|^2L \right)$ &  $\mathcal{O}\left ( |\gR||\gV|L \right)$ \\
      $L$-layer \methodname\!\! & $\mathcal{O}\left ( |\gV|^2L \right)$ & $\mathcal{O}\left( |\gR||\gV|L \right)$ \\
      \midrule
      \multicolumn{3}{c}{\textit{$\mathcal{G}$ is a sparse graph with maximum node degree $\Delta \ll |\gV|$}} \\
      \midrule
      $L$-hop KagNet & \!\!$\mathcal{O}\left(|\gR|^L|\gV|L\Delta^{L} \right)$\!\! &  \!\!$\mathcal{O}\left(|\gR|^L|\gV|L\Delta^{L} \right)$\!\! \\
      $L$-hop MHGRN & $\mathcal{O}\left(|\gR|^2|\gV| L\Delta \right)$ &  $\mathcal{O}\left ( |\gR||\gV|L \right)$ \\
      $L$-layer \methodname\!\! & $\mathcal{O}\left(|\gV|L\Delta \right)$ & $\mathcal{O}\left ( |\gR||\gV|L \right)$ \\
    \bottomrule
    \end{tabular}
    }
    \caption{\textbf{Computation complexity} of different $L$-hop reasoning models on a dense \!/\! sparse graph $\gG=(\gV,\gE)$ with the relation set $\gR$.
    } 
    \label{tab:complexity}
\end{table}

\subsection{Why GNN for question answering?}
\label{sec:method-why-gnn}
We provide more discussion on why we use a GNN for solving question answering and reasoning tasks.

Recent work shows that GNNs are effective for modeling various graph algorithms \cite{xu2019can}. Examples of graph algorithms include knowledge graph reasoning, such as execution of logical queries on a KG \cite{Gentner1983, ren2020beta}:
\begin{align*}
    V_?. ~ \exists V\!: ~&\text{Located}(\text{Europe}, V) \\
    \wedge & \neg \text{Held}(\text{World Cup}, V) \wedge \text{President}(V, V_?)
\end{align*}
\begin{align*}
\textit{(``Who are the presents of European countries}\\[-0.8mm]
\textit{that have \textbf{not} held the World Cup?'')}
\end{align*}
Viewing such logical queries as input ``questions'', we conducted
a pilot study where we apply \methodname to learn the task of executing logical queries on a KG---including complex queries that contain negation or multi-hop relations about entities.
In this task, we find that \methodname significantly outperforms a baseline model that only uses an LM but not a GNN:

\begin{table}[h]
\centering
\small
\scalebox{0.8}{
\begin{tabular}{lc}
\toprule
\textbf{Methods}          & \textbf{Hit@3 on FB15k}     \\
\midrule
LM-only    & 15 \\
\methodname (\textbf{Ours})  &  \textbf{40}  \\
\bottomrule
\end{tabular}
}\vspace{-2mm}
\caption{
{Performace in learning to answer complex logical queries on a KG}.
}
\label{tab:logical_query}
\end{table}

\noindent The result confirms that GNNs are indeed useful for modeling complex query answering.
This provides an intuition that \methodname can be useful for answering complex natural language questions too, which could be viewed as executing soft queries---natural language instead of logical---using a KG.

From this ``KG query execution'' intuition, we may also draw an interpretation that the KG and GNN can provide a \textit{scaffold} for the model to reason about entities mentioned in the question. We further analyze this idea in \S \ref{sec:experiment-structured-reasoning}.

\section{Experiments}\label{sec:experiment}
\subsection{Datasets}
We evaluate \methodname on three question answering datasets: \textit{CommonsenseQA} \cite{talmor2018commonsenseqa}, \textit{OpenBookQA} \cite{obqa}, and \textit{MedQA-USMLE} \cite{jin2021disease}.

\heading{CommonsenseQA} is a 5-way multiple choice QA task that requires reasoning with commonsense knowledge, containing 12,102 questions.
The test set of CommonsenseQA is not publicly available, and model predictions can only be evaluated once every two weeks via the official leaderboard. Hence, we perform main experiments on the {in-house (IH) data splits} used in \citet{lin2019kagnet}, and also report the score of our final system on the {official test} set.

\heading{OpenBookQA} is a 4-way multiple choice QA task that requires reasoning with elementary science knowledge, containing 5,957 questions. We use the official data splits from \citet{mihaylov2018knowledgeable}. 

\heading{MedQA-USMLE} is a 4-way multiple choice QA task that requires biomedical and clinical knowledge. The questions are originally from practice tests for the United States Medical License Exams (USMLE). The dataset contains 12,723 questions. We use the original data splits from \citet{jin2021disease}.

\subsection{Knowledge graphs}
\label{sec:experiment-kg}
For CommonsenseQA and OpenBookQA, we use \textit{ConceptNet} \cite{speer2016conceptnet}, a general-domain knowledge graph, as our structured knowledge source $\mathcal{G}$. It has 799,273
nodes and 2,487,810 edges in total. Node embeddings are initialized using the entity embeddings prepared by \citet{feng2020scalable}, which applies pre-trained LMs to all triples in ConceptNet and then obtains a pooled representation for each entity.

For MedQA-USMLE, we use a self-constructed knowledge graph that integrates the Disease Database portion of the Unified Medical Language System (UMLS; \citealp{bodenreider2004unified}) and DrugBank \cite{wishart2018drugbank}. The knowledge graph contains 9,958 nodes and 44,561 edges. Node embeddings are initialized using the pooled representations of the entity name from SapBERT (\citealp{liu2020self}).

Given each QA context (question and answer choice), we retrieve the subgraph $\mathcal{G}_\text{sub}$ from $\mathcal{G}$ following the pre-processing step described in \citet{feng2020scalable}, with hop size $k=2$. We then prune $\mathcal{G}_\text{sub}$ to keep the top 200 nodes according to the node relevance score computed in \S \ref{sec:method-contextualization}.
Henceforth, in this section (\S \ref{sec:experiment}) we use the term ``KG'' to refer to $\mathcal{G}_\text{sub}$.

\subsection{Implementation \& training details}
We set the dimension ($D=200$) and number of layers ($L=5$) of our GNN module, with dropout rate
0.2 applied to each layer \cite{srivastava2014dropout}. 
We train the model with the RAdam \cite{liu2019variance} optimizer using two GPUs (GeForce RTX 2080 Ti), which takes $\sim$20 hours.
We set the batch size from \{32, 64, 128, 256\}, learning rate for the LM module from \{5e-6, 1e-5, 2e-5, 3e-5, 5e-5\}, and learning rate for the GNN module from \{2e-4, 5e-4, 1e-3, 2e-3\}. The above hyperparameters are tuned on the development set.

\subsection{Baselines}
\label{sec:experiment-baseline}
\paragraph{Fine-tuned LM.}
To study the role of KGs, we compare with a vanilla fine-tuned LM, which does not use the KG. We use RoBERTa-large \cite{liu2019roberta} for \textit{CommonsenseQA}, and RoBERTa-large and AristoRoBERTa\footnote{OpenBookQA provides an extra corpus of scientific {facts} in a textual form. AristoRoBERTa uses the facts corresponding to each question, prepared by \citet{clark2019f}, as an additional input to the QA context.\vspace{-0mm}} \cite{clark2019f} for \textit{OpenBookQA}.
For \textit{MedQA-USMLE}, we use a state-of-the-art biomedical LM, SapBERT \cite{liu2020self}.

\paragraph{Existing LM+KG models.}
We compare with existing LM+KG methods, which share the same high-level framework as ours but use different modules to reason on the KG in place of \methodname (``yellow box'' in Figure\! \ref{fig:overview}):
(1) Relation Network (RN) \cite{santoro2017simple}, (2) RGCN \cite{schlichtkrull2018modeling}, (3) GconAttn \cite{wang2019improving}, (4) KagNet \cite{lin2019kagnet}, and (5) MHGRN \cite{feng2020scalable}.
(1),(2),(3) are relation-aware GNNs for KGs, and (4),(5) further model paths in KGs. MHGRN is the existing top performance model under this LM+KG framework.
For fair comparison, we use the same LM in all the baselines and our model.
The key differences between \methodname and these are that they do not perform {relevance scoring} or {joint updates} with the QA context (\S \ref{sec:method}).

\subsection{Main results}
\begin{table}[t]
\centering
\scalebox{0.7}{
\begin{tabular}{lcc}
    \toprule  
    {\textbf{Methods}}& \textbf{IHdev-Acc.} (\%) & \textbf{IHtest-Acc.} (\%) \\
    \midrule  
    RoBERTa-large (w/o KG)  & 73.07~($\pm$0.45) & 68.69 ($\pm$0.56) \\
    \midrule
    + {RGCN \scalebox{0.7}{\cite{schlichtkrull2018modeling}}} & 72.69~($\pm$0.19) & 68.41~($\pm$0.66) \\
    + {GconAttn \scalebox{0.7}{\cite{wang2019improving}}}  & 72.61($~\pm$0.39) &
    68.59~($\pm$0.96)\\
    + {KagNet \scalebox{0.7}{\cite{lin2019kagnet}}}  & 73.47~($\pm$0.22) & 69.01~($\pm$0.76) \\
    + {RN} \scalebox{0.7}{\cite{santoro2017simple}}  & 74.57~($\pm$0.91) & 69.08~($\pm$0.21) \\
    + {MHGRN} \scalebox{0.7}{\cite{feng2020scalable}} & 74.45~($\pm$0.10)   & {71.11}~($\pm$0.81)  \\
    \midrule
    + \methodname (\textbf{Ours}) &  \textbf{76.54}~($\pm$0.21)   & \textbf{73.41}~($\pm$0.92)  \\
    \bottomrule 
\end{tabular}
}
\vspace{-2mm}
\caption{\textbf{Performance comparison on \textit{Commonsense \!QA} in-house split} (controlled experiments). 
As the official test is hidden, here we report the in-house Dev (IHdev) and Test (IHtest) accuracy, following the data split of \citet{lin2019kagnet}.
}
\label{tab:csqa_main}
\end{table}

\begin{table}[tb]
\centering
\small
\scalebox{0.9}{
\begin{tabular}{lc}
\toprule
\textbf{Methods}& \textbf{Test}  \\
\midrule
RoBERTa~\cite{liu2019roberta} & 72.1 \\

RoBERTa+FreeLB~\cite{zhu2019freelb} (ensemble) &73.1\\
RoBERTa+HyKAS~\cite{ma2019towards}&73.2\\
RoBERTa+KE (ensemble) & 73.3\\
RoBERTa+KEDGN (ensemble) & 74.4\\
XLNet+GraphReason~\cite{lv2020graph} & 75.3\\
RoBERTa+MHGRN \cite{feng2020scalable} & 75.4  \\
Albert+PG \cite{wang2020connecting} & 75.6  \\
Albert~\cite{lan2019albert} (ensemble) &76.5          \\
UnifiedQA\textsuperscript{*}~\cite{khashabi2020unifiedqa} & \textbf{79.1} \\
\midrule
RoBERTa + \methodname (\textbf{Ours}) & 76.1 \\
\bottomrule
\end{tabular}
}\vspace{-2mm}
\caption{\textbf{Test accuracy on \textit{CommonsenseQA}'s official leaderboard}. The top system, UnifiedQA (11B parameters) is 30x larger than our model.
}
\label{tab:csqa_leaderboard}
\end{table}

\begin{table}[tb]
\centering
\scalebox{0.7}{
\begin{tabular}{lcc}
\toprule
\textbf{Methods}          & \textbf{RoBERTa-large}     & \textbf{AristoRoBERTa}     \\
\midrule
Fine-tuned LMs (w/o KG)          &  64.80~($\pm$2.37)  & 78.40~($\pm$1.64)       \\
\midrule
+ RGCN           & 62.45~($\pm$1.57)   & 74.60~($\pm$2.53)
       \\
+ GconAtten      & 64.75~($\pm$1.48)   & 71.80~($\pm$1.21)
       \\
+ RN             & 65.20~($\pm$1.18)    & 75.35~($\pm$1.39)
       \\
+ MHGRN & 66.85~($\pm$1.19) & 80.6        \\
\midrule
+ \methodname (\textbf{Ours})       &  \textbf{67.80}~($\pm$2.75)   &  \textbf{82.77}~($\pm$1.56)    \\

\bottomrule
\end{tabular}
}\vspace{-2mm}
\caption{\textbf{Test accuracy comparison on \textit{OpenBook \!\!QA}} (controlled experiments). Methods with AristoRoBERTa use the textual evidence by \citet{clark2019f} as an additional input to the QA context.}
\label{tab:obqa_main}
\end{table}

\begin{table}[h]
\centering
\small
\scalebox{0.9}{
\begin{tabular}{lc}
\toprule
\textbf{Methods}         & \textbf{Test}           \\
\midrule
Careful Selection~\cite{banerjee2019careful} & 72.0\\
AristoRoBERTa & 77.8\\
KF + SIR~\cite{banerjee2020knowledge} & 80.0\\
AristoRoBERTa + PG \cite{wang2020connecting}      & 80.2 \\
AristoRoBERTa + MHGRN \cite{feng2020scalable}     & 80.6 \\
Albert + KB & 81.0\\
T5\textsuperscript{*}~\cite{t5} & 83.2\\
UnifiedQA\textsuperscript{*}~\cite{khashabi2020unifiedqa} &$\textbf{87.2}$          \\
\midrule
AristoRoBERTa + \methodname (\textbf{Ours}) & 82.8\\
\bottomrule
\end{tabular}
}\vspace{-2mm}
\caption{\textbf{Test accuracy on \textit{OpenBookQA} leaderboard}. All listed methods use the provided science facts as an additional input to the language context. The top 2 systems, UnifiedQA (11B params) and T5 (3B params) are  30x and 8x larger than our model.}
\label{tab:obqa_leaderboard}
\end{table}
\begin{table}
\centering
\small
\scalebox{0.9}{
\begin{tabular}{lc}
\toprule
\textbf{Methods}          & \textbf{Test}     \\
\midrule
{BERT-base}~ \scalebox{0.9}{\cite{devlin2018bert}}       & 34.3 \\ 
{BioBERT-base}~ \scalebox{0.9}{\cite{lee2020biobert}}     & 34.1 \\ 
{RoBERTa-large}~ \scalebox{0.9}{\cite{liu2019roberta}}    & 35.0 \\ 
{BioBERT-large}~ \scalebox{0.9}{\cite{lee2020biobert}}       & 36.7 \\
{SapBERT}~ \scalebox{0.9}{\cite{liu2020self}}     & 37.2 \\
\midrule
{SapBERT + \methodname}  (\textbf{Ours})  &  \textbf{38.0}  \\
\bottomrule
\end{tabular}
}\vspace{-2mm}
\caption{
\textbf{Test accuracy on \textit{MedQA-USMLE}}.
}
\label{tab:medqa_main}
\end{table}

Table \ref{tab:csqa_main} and Table \ref{tab:obqa_main} show the results on CommonsenseQA and OpenBookQA, respectively. On both datasets, we observe consistent improvements over fine-tuned LMs and existing LM+KG models, \eg, on CommonsenseQA, +4.7\% over RoBERTa, and +2.3\% over the prior best LM+KG system, MHGRN. 
The boost over MHGRN suggests that \methodname makes a better use of KGs to perform joint reasoning than existing LM+KG methods.

We also achieve competitive results to other systems on the official leaderboards (Table  \ref{tab:csqa_leaderboard} and \ref{tab:obqa_leaderboard}). Notably, the top two systems, T5 \cite{t5} and UnifiedQA \cite{khashabi2020unifiedqa}, are trained with more data and use 8x to 30x more parameters than our model (ours has $\sim$360M parameters).
Excluding these and ensemble systems, our model is comparable in size and amount of data to other systems, and achieves the top performance on the two datasets.

Table \ref{tab:medqa_main} shows the result on MedQA-USMLE. QA-GNN outperforms state-of-the-art fine-tuned LMs (\eg, SapBERT). This result suggests that our method is an effective augmentation of LMs and KGs across different domains (\ie, the biomedical domain besides the commonsense domain).

\subsection{Analysis}
\subsubsection{Ablation studies}
\label{sec:experiment-ablation}
\begin{table}[!t]
    \hspace{-2mm}
    \includegraphics[width=0.51\textwidth]{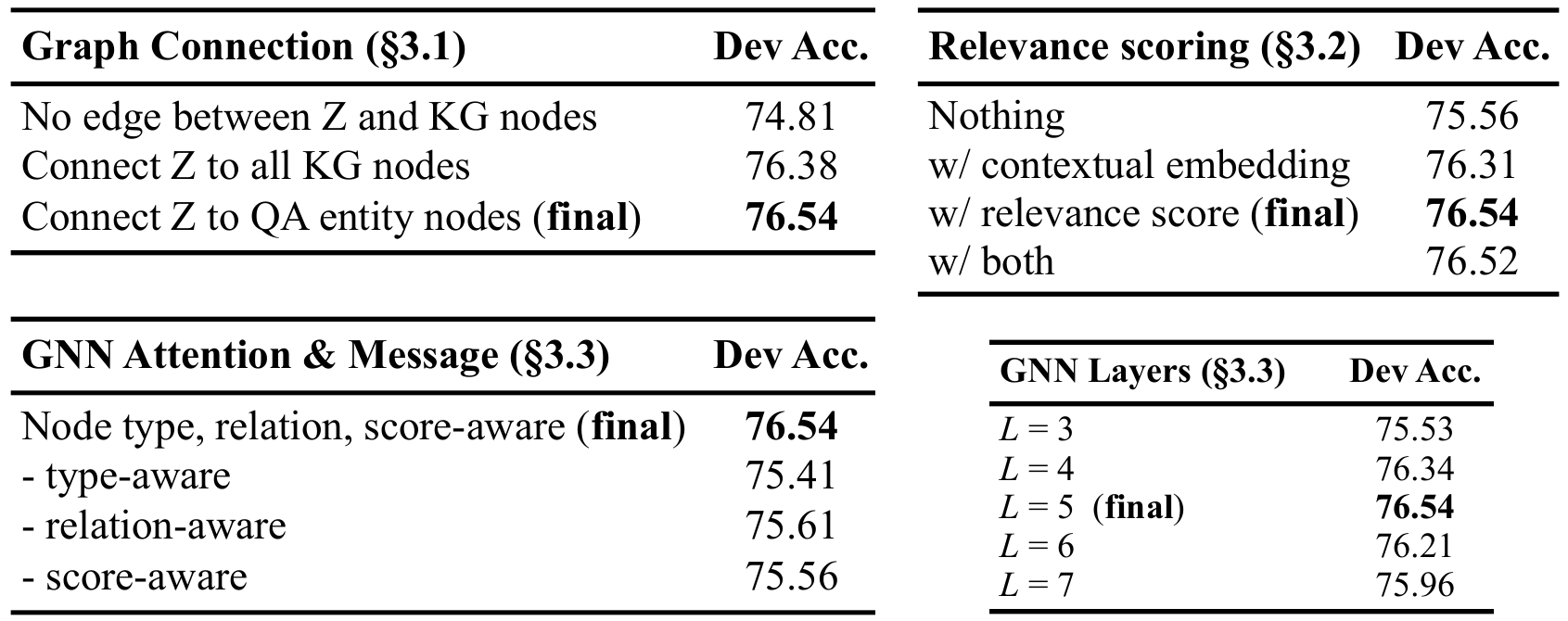}
    \vspace{-7mm}
    \caption{\textbf{Ablation study} of our model components, using the CommonsenseQA IHdev set.}
\label{tbl:ablation}
\end{table}

Table \ref{tbl:ablation} summarizes the ablation study conducted on each of our model components (\S \ref{sec:method-joint-graph}, \S \ref{sec:method-contextualization}, \S \ref{sec:method-gnn}), using the CommonsenseQA IHdev set.

\paragraph{Graph connection}\!\!\!\!(top left table):~~ The first key component of \methodname is the joint graph that connects the $z$ node (QA context) to QA entity nodes $\mathcal{V}_{q,a}$ in the KG (\S \ref{sec:method-joint-graph}). Without these edges, the QA context and KG cannot mutually update their representations, hurting the performance: 76.5\% \!$\rightarrow$\! 74.8\%, which is close to the previous LM+KG system, MHGRN. 
If we connected $z$ to all the nodes in the KG (not just QA entities), the performance is comparable or drops slightly (-0.16\%).

\paragraph{KG node relevance scoring}\!\!\!\!(top right table):~~
We find the relevance scoring of KG nodes (\S \ref{sec:method-contextualization}) provides a boost: 75.56\% $\!\rightarrow\!$ 76.54\%.
As a variant of the relevance scoring in Eq.\! \ref{eq:relevance}, we also experimented with obtaining a \emph{contextual embedding} $\vw_v$ for each node $v\in\snode$ and adding to the node features:
$
\vw_v=\fembed([\nodetext{z};~ \nodetext{v}])
$.
However, we find that it does not perform as well (76.31\%), and
using both the relevance score and contextual embedding performs on par with using the score alone, suggesting that the score has a sufficient information in our tasks; hence, our final system simply uses the relevance score.

\paragraph{GNN architecture}\!\!\!\!(bottom tables):~~ We ablate the information of node type, relation, and relevance score from the attention and message computation in the GNN (\S \ref{sec:method-gnn}). The results suggest that all these features improve the model performance. 
For the number of GNN layers, we find $L=5$ works the best on the dev set. Our intuition is that 5 layers allow various message passing or reasoning patterns between the QA context ($z$) and KG, such as ``$z$ $\!\rightarrow\!$ 3 hops on KG nodes $\!\rightarrow\!$ $z$''.

\subsubsection{Model interpretability}
\begin{figure}[!t]
    \hspace{-1mm}
    \centering 
    \includegraphics[width=0.45\textwidth]{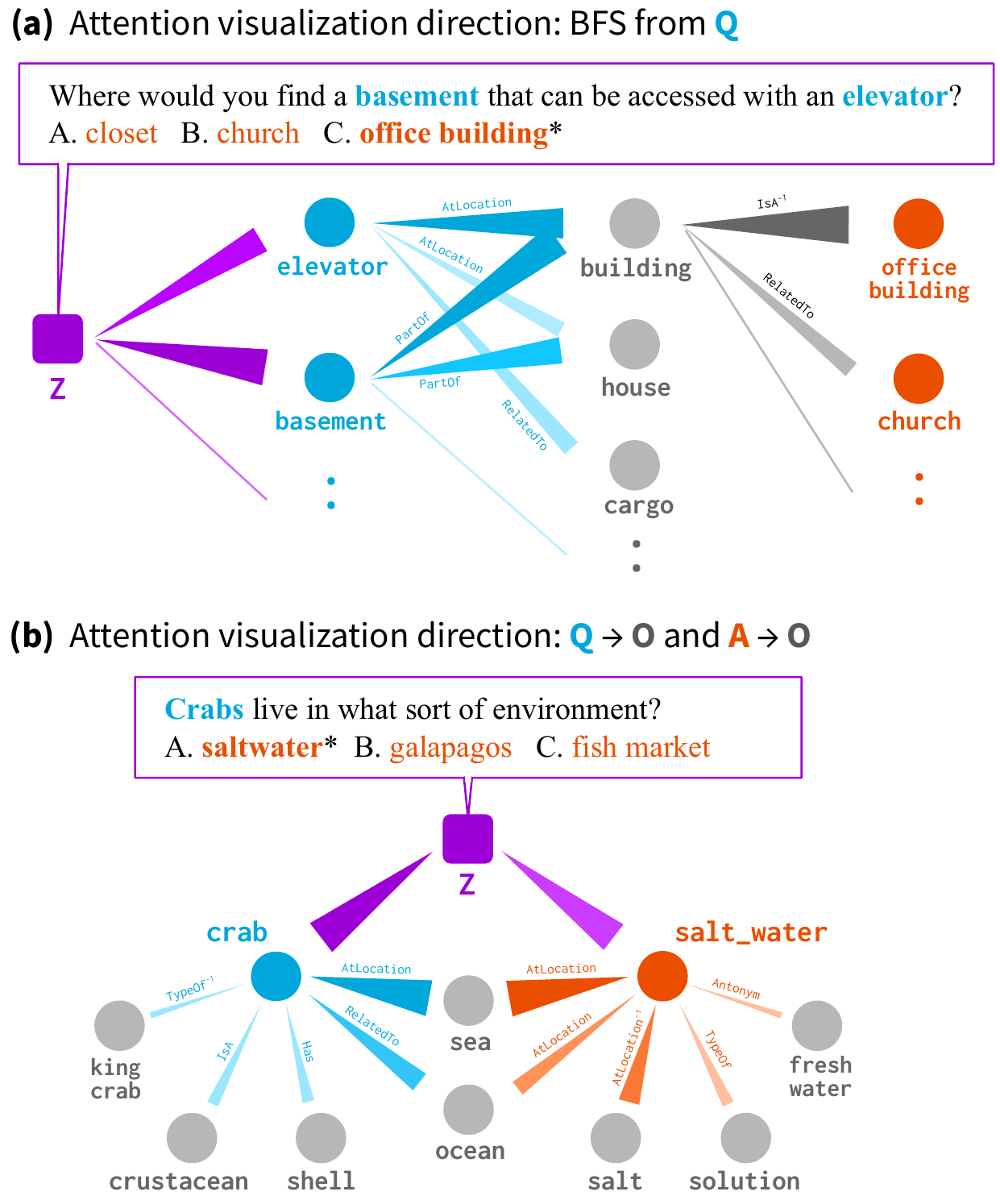}
    \caption{\textbf{Interpreting \methodname's reasoning process} by analyzing the node-to-node attention weights induced by the GNN. Darker and thicker edges indicate higher attention weights.
    }
\label{fig:interpret}
\end{figure}

We aim to interpret \methodname's reasoning process by analyzing the node-to-node attention weights induced by the GNN. 
Figure \ref{fig:interpret} shows two examples. In (a), we perform Best First Search (BFS) on the working graph to trace high attention weights from the QA context node (\textbf{Z}; purple) to \textbf{Q}uestion entity nodes (blue) to \textbf{O}ther (gray) or \textbf{A}nswer choice entity nodes (orange), which reveals that the QA context $z$ attends to ``elevator'' and ``basement'' in the KG, ``elevator'' and ``basement'' both attend strongly to ``building'', and ``building'' attends to ``office building'', which is our final answer. 
In (b), we use BFS to trace attention weights from two directions: 
\textbf{Z} $\!\rightarrow\!$ \textbf{Q} $\!\rightarrow\!$ \textbf{O} and \textbf{Z} $\!\rightarrow\!$ \textbf{A} $\!\rightarrow\!$ \textbf{O}, which reveals concepts (``sea'' and ``ocean'') in the KG that are not necessarily mentioned in the QA context but bridge the reasoning between the question entity (``crab'') and answer choice entity (``salt water'').
While prior KG reasoning models \cite{lin2019kagnet,feng2020scalable} enumerate individual paths in the KG for model interpretation, \methodname is not specific to paths, and helps to find more general reasoning structures (\eg, a KG subgraph with multiple anchor nodes as in example (a)).

\begin{figure*}[!th]
    \vspace{-5mm}
    \hspace{-1mm}
    \centering 
    \includegraphics[width=0.92\textwidth]{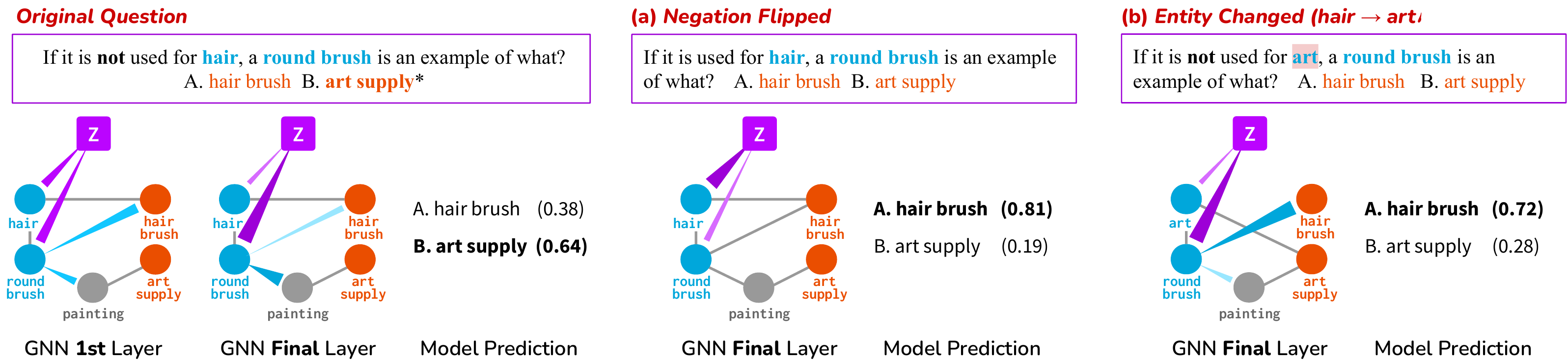}
    \vspace{-1mm}
    \caption{\textbf{Analysis of \methodname's behavior for structured reasoning}. Given an original question (left), we modify its negation (middle) or topic entity (right): we find that \methodname adapts attention weights and final predictions accordingly, suggesting its capability to handle structured reasoning.
    }
    \vspace{-2mm}
\label{fig:structure_reason}
\end{figure*}

\begin{table*}[!th]
    \vspace{-0mm}
    \hspace{-1mm}
    \centering 
    \includegraphics[width=0.92\textwidth]{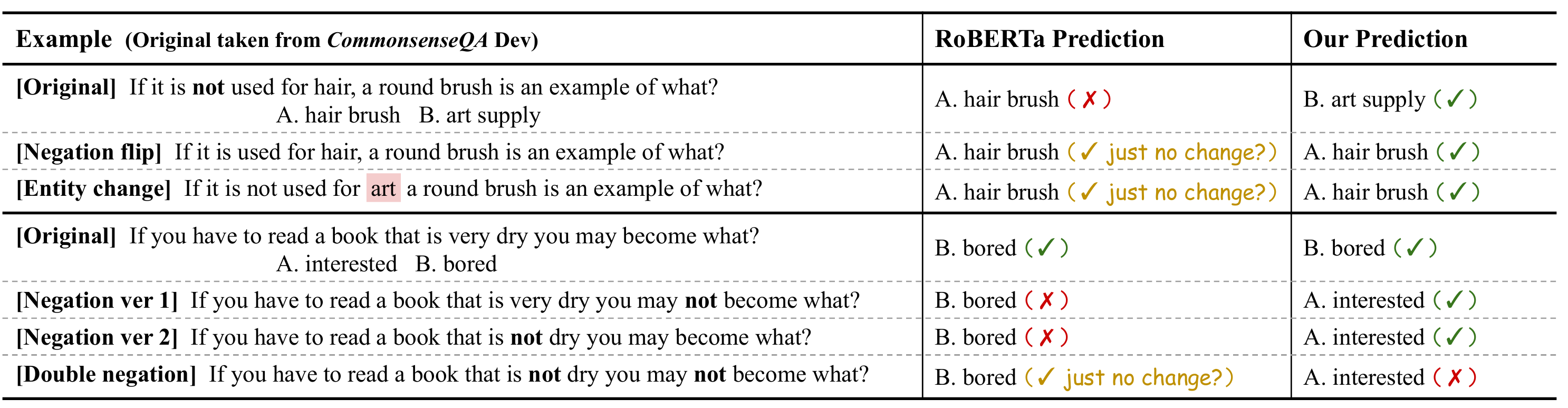}
    \vspace{-2mm}
    \caption{\textbf{Case study of structured reasoning}, comparing predictions by RoBERTa and our model (RoBERTa + \methodname). Our model correctly handles changes in negation and topic entities.
    }
\label{tbl:structure_reason}
\end{table*}

\subsubsection{Structured reasoning}
\label{sec:experiment-structured-reasoning}
Structured reasoning, \eg, precise handling of negation or entity substitution (\eg, ``hair'' $\!\rightarrow\!$ ``art'' in Figure \!\ref{fig:structure_reason}b) in question, is crucial for making robust predictions. Here we analyze \methodname's ability to perform structured reasoning and compare with baselines (fine-tuned LMs and existing LM+KG models).

\paragraph{Quantitative analysis.}
\begin{table}[tb]
\centering
\scalebox{0.7}{
\begin{tabular}{lcc}
\toprule
\multirow{2}{*}{\textbf{Methods}  }        & \textbf{IHtest-Acc.}     & \textbf{IHtest-Acc.}     \\
& (Overall) & \scalebox{0.93}[1]{(Question w/ \textbf{negation})}\! \\
\midrule
RoBERTa-large (w/o KG)          &  68.7 & 54.2     \\
\midrule
+ KagNet &  69.0 (+0.3) &
54.2 (+0.0)
        \\
+ MHGRN        & 71.1 (+2.4) & 54.8 (+0.6) \\
\midrule
+ \methodname (\textbf{Ours})       &  {73.4 (+4.7)}   &  \textbf{58.8 (+4.6)}
     \\[2pt]
\begin{tabular}{@{}l@{}}+ \methodname (no edge\\[-3pt] \hspace{9pt}between Z and KG) \end{tabular}   &   71.5 (+2.8) &
55.1 (+0.9)
     \\
\bottomrule
\end{tabular}\vspace{-3mm}
}
\caption{Performance on \textbf{questions with negation} in \textit{CommonsenseQA}.  () shows the difference with RoBERTa.
Existing LM+KG methods (KagNet, MHGRN) provide limited improvements over RoBERTa (+0.6\%); \methodname exhibits a bigger boost (+4.6\%), suggesting its strength in structured reasoning. 
}
\label{tab:negation_result}
\end{table}

Table \ref{tab:negation_result} compares model performance on questions containing negation words (\eg, no, not, nothing, unlikely), taken from the CommonsenseQA IHtest set.
We find that previous LM+KG models (KagNet, MHGRN) provide limited improvements over RoBERTa on questions with negation (+0.6\%); whereas \methodname exhibits a bigger boost (+4.6\%), suggesting its strength in structured reasoning.
We hypothesize that \methodname's joint updates of the representations of the QA context and KG (during GNN message passing) allows the model to integrate semantic nuances expressed in language.
To further study this hypothesis, we remove the connections between $z$ and KG nodes from our \methodname (Table \ref{tab:negation_result} bottom): now the performance on negation becomes close to the prior work, MHGRN, suggesting that the joint message passing helps for performing structured reasoning.

\paragraph{Qualitative analysis.}

Figure \ref{fig:structure_reason} shows a case study to analyze our model's behavior for structured reasoning. 
The question on the left contains negation ``\textbf{not} used for hair'', and the correct answer is ``B. art supply''.
We observe that in the 1st layer of QA-GNN, the attention from $z$ to question entities (``hair'', ``round brush'') is diffuse. After multiples rounds of message passing on the working graph, $z$ attends strongly to ``round brush'' in the final layer of the GNN, but weakly to the negated entity ``hair''. The model correctly predicts the answer ``B. art supply''.
Next, given the original question on the left, we (a) drop the negation or (b) modify the topic entity (``hair'' $\!\rightarrow\!$ ``art'').
In (a), $z$ now attends strongly to ``hair'', which is not negated anymore. The model predicts the correct answer ``A. hair brush''.
In (b), we observe that QA-GNN recognizes the same structure as the original question (with only the entity swapped): $z$ attends weakly to the negated entity (``art'') like before, and the model correctly predicts ``A. hair brush'' over ``B. art supply''.

Table \ref{tbl:structure_reason} shows additional examples, where we compare QA-GNN's predictions with the LM baseline (RoBERTa).
We observe that RoBERTa tends to make the same prediction despite the modifications we make to the original questions (\eg, drop/insert negation, change an entity); on the other hand, \methodname adapts predictions to the modifications correctly (except for double negation in the table bottom, which is a future work).

\subsubsection{Effect of KG node relevance scoring}
\begin{table}[tb]
\centering
\scalebox{0.7}{
\begin{tabular}{lcc}
\toprule
\multirow{2}{*}{\vspace{-11pt}\textbf{Methods}  }        & \textbf{IHtest-Acc.}     & \textbf{IHtest-Acc.}     \\
& \begin{tabular}{@{}c@{}}(Question w/ \\[-2pt] \hspace{6pt}\scalebox{0.9}{$\leq$}10 entities) \end{tabular}
&  \begin{tabular}{@{}c@{}}(Question w/ \\[-2pt] \hspace{6pt}\scalebox{0.9}{$>$}10 entities) \end{tabular}\\
\midrule
RoBERTa-large (w/o KG)          &   68.4 &
70.0\\
\midrule
+ MHGRN        &  71.5& 70.1
\\
\midrule
\begin{tabular}{@{}l@{}}+ \methodname (w/o node\\[-3pt] \hspace{9pt}relevance score) \end{tabular}       &   72.8 (+1.3) &
71.5 (+1.4)\\[5pt]
\begin{tabular}{@{}l@{}}+ \methodname (w/ node\\[-3pt] \hspace{9pt}relevance score; \textbf{final system})\end{tabular}\!\!   &    73.4 (+1.9) &
\textbf{73.5 (+3.4)}\\
\bottomrule
\end{tabular}\vspace{-3mm}
}
\caption{Performance on \textbf{questions with fewer \!/\! more entities} in \textit{CommonsenseQA}.  () shows the difference with MHGRN (LM+KG baseline). KG node relevance scoring (\S 3.2) boosts the performance on questions containing more entities (i.e. larger retrieved KG). 
}
\label{tab:kg_size_result}
\end{table}
We find that KG node relevance scoring (\S \ref{sec:method-contextualization}) is helpful when the retrieved KG ($\mathcal{G}_\text{sub}$) is large.
Table \ref{tab:kg_size_result} shows model performance on questions containing fewer ($\leq$10) or more (>10) entities in the CommonsenseQA IHtest set (on average, the former and latter result in 90 and 160 nodes in $\mathcal{G}_\text{sub}$, respectively).
Existing LM+KG models such as MHGRN achieve limited performance on questions with more entities due to the size and noisiness of retrieved KGs: 70.1\% accuracy vs 71.5\% accuracy on questions with fewer entities. 
KG node relevance scoring mitigates this bottleneck, reducing the accuracy discrepancy: 73.5\% and 73.4\% accuracy on questions with more \!/\! fewer entities, respectively.

\section{Related work and discussion}\label{sec:related}

\paragraph{Knowledge-aware methods for NLP.}
Various works have studied methods to augment natural language processing (NLP) systems with knowledge.
Existing works \cite{pan2019improving,ye2019align,petroni2019language,Bosselut2019COMETCT} study pre-trained LMs' potential as latent knowledge bases. To provide more explicit and interpretable knowledge, several works integrate structured knowledge (KGs) into LMs 
\cite{mihaylov2018knowledgeable,lin2019kagnet,wang2019improving,yang2019enhancing,wang2020connecting,bosselut2021dynamic}.

\paragraph{Question answering with LM+KG.}
In particular, a line of works propose LM+KG methods for question answering. Most closely related to ours are works by \citet{lin2019kagnet,feng2020scalable,lv2020graph}. Our novelties are (1) the joint graph of QA context and KG, on which we \textit{mutually} update the representations of the LM and KG; and (2) \textit{language-conditioned} KG node relevance scoring.
Other works on scoring or pruning KG nodes/paths rely on graph-based metrics such as PageRank, centrality, and off-the-shelf KG embeddings \cite{paul2019ranking,fadnis2019heuristics,bauer2018commonsense,lin2019kagnet}, without reflecting the QA context.

\paragraph{Other QA tasks.}
Several works study other forms of question answering tasks, \eg, passage-based QA, where systems identify answers using given or retrieved documents \cite{rajpurkar2016squad,joshi2017triviaqa,yang2018hotpotqa},
and 
KBQA, where systems perform semantic parsing of a given question and execute the parsed queries on knowledge bases \citep{berant2013semantic,yih2016value,yu2018spider}.
Different from these tasks, we approach question answering using knowledge available in LMs and KGs.

\paragraph{Knowledge representations.}
Several works study joint representations of external textual knowledge (\eg, Wikipedia articles) and structured knowledge (\eg, KGs) \cite{riedel2013relation,toutanova2015representing,xiong2019improving,sun2019pullnet,wang2019kepler}.
The primary distinction of our joint graph representation is that we construct a graph 
connecting each \textit{question} and KG rather than textual and structural knowledge, approaching a complementary problem to the above works.

\paragraph{Graph neural networks (GNNs).}
GNNs have been shown to be effective for modeling graph-based data.
Several works use GNNs to model the structure of text \cite{yasunaga2017graph,zhang2018graph,yasunaga2020repair} or KGs \cite{Wang2020Entity}. In contrast to these works, \methodname jointly models the language and KG.
Graph Attention Networks (GATs) \cite{velikovi2017graph} perform attention-based message passing to induce graph representations. 
We build on this framework, and further {condition} the GNN on the language input by introducing a QA context node (\S 3.1), KG node relevance scoring (\S 3.2), and joint update of the KG and language representations (\S 3.3).

\section{Conclusion}\label{sec:conclusion}
We presented \methodname, an end-to-end question answering model that leverages LMs and KGs. 
Our key innovations include (i) \textbf{Relevance scoring}, where we compute the relevance of KG nodes conditioned on the given QA context,
and (ii) \textbf{Joint reasoning} over the QA context and KGs, where we connect the two sources of information via the working graph, and jointly update their representations through GNN message passing.
Through both quantitative and qualitative analyses, we showed \methodname's improvements over existing LM and LM+KG models on question answering tasks, as well as its capability to perform interpretable and structured reasoning, \eg, correctly handling negation in questions.

\section*{Acknowledgment}
We thank Rok Sosic, Weihua Hu, Jing Huang, Michele Catasta, members of the Stanford SNAP, P-Lambda and NLP groups and Project MOWGLI team, as well as our anonymous reviewers for valuable feedback.
We gratefully acknowledge the support of DARPA under Nos. N660011924033 (MCS); Funai Foundation Fellowship; ARO under Nos. W911NF-16-1-0342 (MURI), W911NF-16-1-0171 (DURIP); NSF under Nos. OAC-1835598 (CINES), OAC-1934578 (HDR), CCF-1918940 (Expeditions), IIS-2030477 (RAPID); Stanford Data Science Initiative, Wu Tsai Neuro-sciences Institute, Chan Zuckerberg Biohub, Amazon, JP-Morgan Chase, Docomo, Hitachi, JD.com, KDDI, NVIDIA, Dell, Toshiba, and United Health Group. Hongyu Ren is supported by Masason Foundation Fellowship and the Apple PhD Fellowship. Jure Leskovec is a Chan Zuckerberg Biohub investigator.

\section*{Reproducibility}
\renewcommand\ttdefault{cmtt}
Code and data are available at\\
\url{https://github.com/michiyasunaga/qagnn}.\\
Experiments are available at\\
\url{https://worksheets.codalab.org/worksheets/0xf215deb05edf44a2ac353c711f52a25f}.

\bibliography{main}
\bibliographystyle{acl_natbib}

\end{document}